\documentclass[sigconf]{aamas} 

\usepackage[mode=buildnew]{standalone}
\usepackage{tikz}
\usetikzlibrary{positioning, arrows, calc, shapes.multipart}
\usetikzlibrary{decorations.pathreplacing}
\usepackage{multirow}
\usepackage{algorithmic}
\usepackage{algorithm}
\usepackage{graphicx}
\usepackage{amsmath}
\usepackage{mathptmx}
\usepackage{xspace}
\usepackage{enumitem}
\usepackage[arrowdel]{physics}
\usepackage{tabularx} 
\newcommand{\bigsum}{\mathop{\raisebox{-0.5ex}{\Huge $\displaystyle \Sigma$}}}

\usepackage{hyperref} 
\usepackage{footnote} 

\pdfoutput=1 

\acmConference[AAMAS '24]{This is a supplementary material for the extended abstract in \cite{10.5555/3635637.3663131}}
\acmDOI{}
\acmPrice{}
\acmISBN{}

\acmSubmissionID{262}

\title[ACM Reference Format]{Combinatorial Client-Master Multiagent Deep Reinforcement Learning for Task Offloading in Mobile Edge Computing}

\author{Tesfay Zemuy Gebrekidan}
\affiliation{
  \institution{University of Southampton}
  \city{Southampton}
  \country{United Kingdom}}

\author{Sebastian Stein}
\affiliation{
  \institution{University of Southampton}
  \city{Southampton}
  \country{United Kingdom}}

\author{Timothy J.Norman}
\affiliation{
  \institution{University of Southampton}
  \city{Southampton}
  \country{United Kingdom}}

\begin{abstract}
Recently, there has been an explosion of mobile applications that perform computationally intensive tasks such as video streaming, data mining, virtual reality, augmented reality, image processing, video processing, face recognition, and online gaming. However, user devices (UDs), such as tablets and smartphones, have a limited ability to perform the computation needs of the tasks. Mobile edge computing (MEC) has emerged as a promising technology to meet the increasing computing demands of UDs. Task offloading in MEC is a strategy that meets the demands of UDs by distributing tasks between UDs and MEC servers. Deep reinforcement learning (DRL) is gaining attention in task-offloading problems because it can adapt to dynamic changes and minimize online computational complexity. However, the various types of continuous and discrete resource constraints on UDs and MEC servers pose challenges to the design of an efficient DRL-based task-offloading strategy. Existing DRL-based task-offloading algorithms focus on the constraints of the UDs, assuming the availability of enough storage resources on the server. Moreover, existing multiagent DRL (MADRL)--based task-offloading algorithms are homogeneous agents and consider homogeneous constraints as a penalty in their reward function. We proposed a novel combinatorial client-master MADRL (CCM\_MADRL) algorithm for task offloading in MEC (CCM\_MADRL\_MEC) that enables UDs to decide their resource requirements and the server to make a combinatorial decision based on the requirements of the UDs. CCM\_MADRL\_MEC is the first MADRL in task offloading to consider server storage capacity in addition to the constraints in the UDs. By taking advantage of combinatorial action selection, CCM\_MADRL\_MEC has shown superior convergence over existing benchmark and heuristic algorithms. 

\end{abstract}


\newcommand{\BibTeX}{\rm B\kern-.05em{\sc i\kern-.025em b}\kern-.08em\TeX}

\begin{document}


\pagestyle{fancy}
\fancyhead{}


\maketitle 


\section{Introduction}

Recently, there has been an explosion of mobile applications that perform computation-intensive tasks, such as video streaming, virtual reality, augmented reality, image processing, video processing, face recognition, and online gaming~\cite{6616113, 10004609, 9943689, 8372737}. However, UDs, such as tablets and smartphones, have limited ability to perform the computation tasks of these applications. Mobile Cloud Computing (MCC) has been considered the key technology to meet the computing needs of UDs by offloading their computation tasks to the cloud~\cite{MAHENGE20221048}. One of the challenges of MCC is latency caused by the distance of the MCC server from the UDs~\cite{sajnani2018latency}. MEC has emerged as a promising technology for addressing the challenges of MCC and the increasing computing demands of UDs by providing MCC services on the edge of the network.

Task offloading in MEC has become an attractive solution to meet the diverse computing needs of UDs~\cite{ISLAM2021102225} by distributing computational tasks between UDs and MEC servers. Many existing task-offloading algorithms use traditional convex optimization methods for single-agent offloading scenarios~\cite{SADATDIYNOV2023450}. DRL is a common solution for task offloading problems due to its advantage in reducing online computational complexity~\cite{nie2023deep} and adapting to dynamic changes~\cite{liu2022novel}. However, the existence of various types of resource constraints on UDs and MEC servers and the combination of discrete, continuous, and combinatorial action spaces pose challenges to the design of an efficient DRL-based task-offloading strategy. UDs have limitations such as finite battery life and limited computational capabilities~\cite{8372737, 8647523}, as well as quality of service (QoS) requirements such as latency. Similarly, MEC servers come with storage constraints. DRL techniques, such as deep Q network (DQN), have yielded encouraging results by modeling the task-offloading problem as MDP with deep neural network (DNN) for function approximation~\cite{app122111260}. However, due to the curse of dimensionality, DQN is insufficient for learning with large discrete action spaces~\cite{dulac2015deep} and a combination of continuous and discrete action spaces~\cite{9110595}. Although multiagent deep deterministic policy gradient (MADDPG) algorithms can handle continuous action spaces, the representation of discrete and continuous action spaces still poses a challenge~\cite{9110595, 9829326}. Despite the advances of MADRL in task offloading, such as cooperative offloading decisions~\cite{9573404} and mixed continuous and discrete action spaces~\cite{9110595, 9829326}, most existing MADRL-based algorithms still formulate the constraints as a penalty in their reward function. 

A comprehensive survey on task offloading by \cite{ISLAM2021102225} has presented task offloading strategies in MEC from different perspectives, including the computational model, the decision-making entity, and the algorithm paradigm. Many algorithms have considered the wireless communication resource and the computing resource of the server. For example, the insufficient computing resource of the MEC server can be alleviated by using MEC-MCC collaboration or collaboration among multiple MEC servers~\cite{10004609}. Sub-channels are considered in the state and action spaces by some DRL-based task offloading algorithms~\cite{9573404, 9829326}. However, the storage constraint on the server is overlooked in existing DRl-based task-offloading algorithms. 

The main contributions of this work are fourfold. 
\begin{itemize}
    \item We proposed a novel CCM\_MADRL algorithm for task offloading in MEC with various types of constraints at the UDs, the wireless network, and the server. Client agents are deployed at the UDs to decide their resource allocation, and a master agent is deployed at the server to make combinatorial decisions based on the actions of the clients. The constraints of the UDs are considered as a penalty in the reward of the client agents, whereas the channel and storage constraints are considered in the combinatorial decision of the master agent. 
    \item By avoiding the number of sub-channels from the state and action spaces, and considering it as a constraint in the combinatorial action selection, we reduced the dimensionality
    \item This is the first DRL-based task offloading algorithm to consider combinations of continuous and discrete resource constraints on the UDs, the communication channel, and the storage capacity of the server.
    \item We develop different heuristic benchmarking methodologies and perform numerical analysis to determine the efficacy of the proposed algorithm.
\end{itemize}

\section{System Model}

\label{CC:sytemmodle}

\begin{figure}[t]
\centerline{\includegraphics[width=1\linewidth]{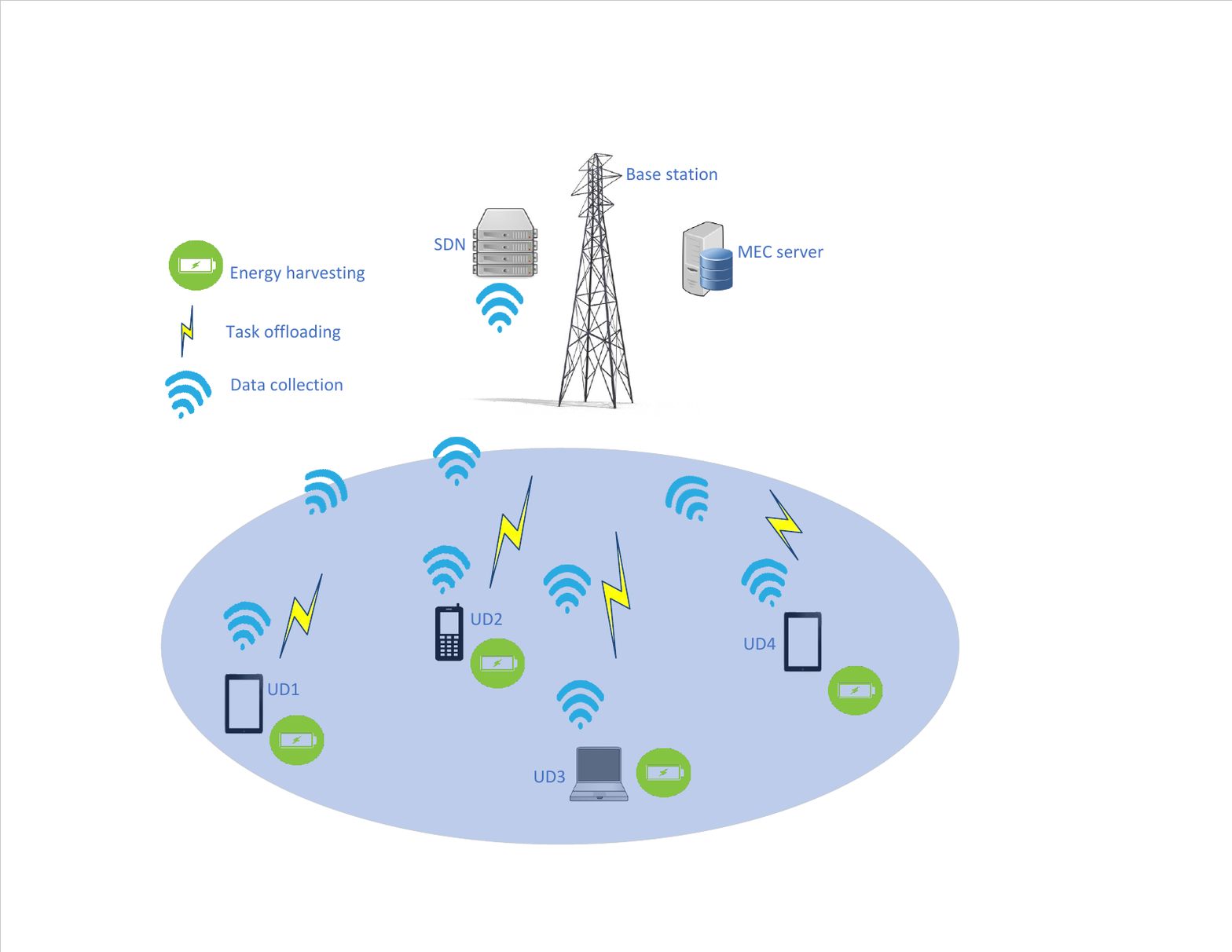}}
\caption{Network model}
\label{CCMMADRLMECmodel}
\end{figure}

This section considers MEC for task offloading, which mainly includes a base station (BS), UDs, tasks, energy harvesting, and wireless networks. We consider a multi-user MEC scenario shown in Figure~\ref{CCMMADRLMECmodel}. In this scenario, there is a single wireless BS equipped with an MEC server that provides a computing and storage service and an SDN controller that controls communication between UDs and BS. The BS serves a set of  $N$ = \{1, 2, 3, \dots, N\} UDs. A UD in the set $N$ is indicated by $n$. For local processing, we consider that each UD $n$ has a minimum and maximum computational resource allocation budgets denoted by $f_n^{min}$ and $f_n^{max}$, respectively, in Gigahertz (GHz) cycles per second. Similarly, to offload its task to the server, we consider that each UD has a minimum and maximum transmission power allocation threshold denoted by $p_n^{min}$ and $p_n^{max}$, respectively, in dBm. Furthermore, we consider a UD to have a minimum battery threshold of $b_n^{min}$ and a maximum battery capacity of $b_n^{max}$ in Megajoules (MJ). The BS has multiple constraints and characteristics, such as the server storage constraint $z_e$ in bits and the number of processing units on the server $U_e$ each having an equal processing capacity of $f_e$ in gigahertz cycles per second. Similar to \cite{9573404, 9829326}, we consider a wireless network of bandwidth of $W$ in megahertz that is equally divided between $K$ sub-channels. The list of notation and terms used in this work is presented in Table~\ref{notationsofccmmadrlmec}.

We consider a task-offloading problem for $T$ time steps of $\tau_{max}$ length each. It is assumed that each UD $n$ generates one task at each time step. If the processing of a task is not completed in $\tau_{max}$, it is discarded before the next time step starts. The task model, the processing model, and the energy harvesting are described in the following sections.

The task offloading model is designed by combining the settings of different existing works. Since \cite{9573404} has used a data set from Huawei Technologies, we adapted it disregarding the blockchain part. We considered the energy harvesting process in \cite{9110595}, and took advantage of the efficient estimation of the completion times of tasks by \cite{xiong2023multi}, which computes the completion time of tasks on the server based on the completion time of other tasks scheduled before them. Similarly, we assume that the server processes the tasks in the order of their arrival on the server. The arrival times of the tasks are determined by their offloading times. In the following, we present the task and computing model for the task offloading problem. 

\subsection{Task Model}

The task model is based on the setting by \cite{9573404} on a data set from Huawei Telecom. At each time step, each UD $n$ generates a task denoted by its notation as $n$\footnote{Because a UD has one task at a time step, we use $n$ to denote both the UD and its task to reduce the number of notations} that is represented by characteristics such as the size of the task $z_n$ in bytes, the number of CPU cycles per bit required to process the task $c_n$, and the maximum deadline $\tau_n$ to which task processing is expected to finish.

Before processing the task, there are three decision variables: a binary decision of whether to process it locally or offload it to the MEC server $x_n$, a local resource allocation $f_n$, and a transmission power allocation $p_n$, which are described in detail in Section~\ref{Bothactions}. So, we make a binary decision $X = \{x_n | n \in N \}$ to describe the processing mode, as seen in Equation~(\ref{cc:x_n}).
\begin{equation}
\label{cc:x_n}
x_{n} = 
\begin{cases}
1, & \text{MEC processing}\\
0, & \text{Local processing}
\end{cases}
\end{equation}
Next, we present the local and MEC models.

\begin{table}[htbp]
\begin{center}
\begin{tabular}{|c|c|}
\hline
\textbf{Notation} & \textbf{Description} \\
\hline
$N$  & Set of UDs\\
$T$  & Number of time steps\\
$\tau_{max}$ & Maximum length of a time step\\
$n$ & A UD or a task of the UD \\
$T_n$ & Total latency of processing task $n$\\
$x_n$ & Binary indicator of local or offload for task $n$\\
$E_n$ & Energy consumption of task $n$\\
$E_{\text{loc}_n}$ & Energy consumption of task $n$ in local processing\\
$E_{\text{off}_n}$ & Energy consumption of offloading task $n$ \\
$T_{\text{loc}_n}$ & Computation time for the local processing of task $n$ \\
$T_{\text{off}_n}$ &  Offloading time of task $n$ \\ 
$\lambda_1$ \& $\lambda_2$ & weight coefficients of $T_n$ and $E_n$\\
$b_n$ & Battery level of US $n$\\
$p_n$ & Transmission power allocation of the UD $n$\\
$z_n$ & Size of task $n$ \\
$z_e$ & Storage capacity of the server\\
$U_e$ & Number of processing units in the server\\
$L_n$ & Cost of processing task $n$\\
$R_n$ & Reward of processing task $n$\\
$f_{n}$ & Resource allocation for local processing of task $n$\\
$C_{n}$ & Number of CPU cycles to process one bit of task $n$\\
$\alpha_\phi$ & Learning rage of the master agent \\
$\alpha_{\theta}$ & Learning rage of client agent \\
$J$ & Joules \\
\hline
\end{tabular}
\caption{List of notations and terms}
\label{notationsofccmmadrlmec}
\end{center}
\end{table}
\subsection{Local Processing}
A task is processed locally if one of the following two conditions happens as presented in Section~\ref{Bothactions}: if the UD decides to process the task locally; if a UD proposes the task to the master agent but the master agent did not select it to be offloaded in the combinatorial action selection. Then, the UD processes the task using its local computational resource assigned to its task, which is restricted within its own resource allocation budget as $f_{n} | f_n^{min} \leq f_{n} \leq f_n^{max}$. The local computing latency to process the task is computed as $T_{\text{loc}_n} = \frac{z_n c_n}{f_{n}}$. The energy consumption in the local processing mode is calculated based on the size of the task and the allocation of resources to process the task, as shown in Equation~(\ref{Eloc}).
\begin{equation}
    E_{\text{loc}_n} = \kappa z_n c_n (f_{n})^2
    \label{Eloc}
\end{equation} where $\kappa$ is energy consumption coefficient~\cite{9110595}.

\subsection{MEC Processing}
In this mode, the task is transferred to the MEC server to be processed by one of the processing units $U_e$ of the server. The decision happens when the UD proposes the task to be offloaded and the master agent accepts it. To be processed on the server, the task needs transmission resources, which are a function of transmission power $p_n$. The transmission power $p_n$ is decided by the UD from its transmission power budget $p_{n} | p_n^{min} \leq p_{n} \leq p_n^{max}$ as discussed in Section~\ref{Bothactions}. Then, the data transmission rate $d_n$ for a single channel of the wireless network is calculated using Shannon's capacity as
\begin{equation}
\label{transmitioncapacity}
    d_n = \frac {W}{K} \log_2\left (1 \text{ + } p_{n} g_{n}\right)
\end{equation} where $g_n$ = $h_n$/{$\sigma^2$} is the normalized channel gain of the uplink channel between UD $n$ and the BS, with channel gain $h_n$ and the background noise variance $\sigma^2$. The channel gain $h_n$ is impacted by many factors, including distance. For simplicity, we assume that the UDs are stationary and have a stationary normalized channel gain depending on their distance from the BS. The variance of background noise $\sigma^2$ is also constant. We did not consider interference between multiple UDs because we assume that a channel is used by one task at a time. 

Once the data transmission rate is determined, the transmission time $T_{\text{off}_n}$ is computed as:
\begin{equation}
    \label{Tappo}
    T_{\text{off}_n} = \frac{z_n}{d_n}
\end{equation}

Then, the energy consumption of offloading task $n$ to the server is calculated as $E_{\text{off}_n} = p_{n} T_{\text{off}_n}$. Like many works on task offloading~\cite{app12126154, 9573404, 9110595}, we assume that the communication resource required to return the information about the processed task to the UD is negligible. 

Note that the energy consumption in task offloading is computed only for the UDs as they are battery-powered. However, the latency of processing the tasks on the server matters because the tasks have deadline constraints. Therefore, the total latency of processing a task on the server is determined by the transmission time, the earliest availability of the processing unit on the server, and the time required to process the task on the server. The processing time of the task $n$ in one of the processing units on the server is computed as $T_{\text{ser}_n} = \frac{z_n c_n}{f_{e}}$. However, the processing of the task on the server does not start as soon as the task has arrived at the server. The processing units on the server process one task at a time. Tasks transferred to the server are processed in the order of arrival at the server, which is determined by $T_{\text{off}_n}$. The tasks are assigned to the earliest free processing unit. Therefore, the start of processing task $n$ depends on the earliest availability of a processing unit, which is determined by the number of processing units on the server $U_e$, and $T_{\text{ser}_n}$ and $T_{\text{off}_n}$ of other tasks that have shorter $T_{\text{off}_n}$ than that of task $n$. Consequently, the total latency of the offloading task $n$ to the MEC server $T_{\text{MEC}_n}$ is calculated as $T_{\text{MEC}_n} = T_{\text{ser}_n} \text{ + } \max\left(T_{\text{off}_n}, T_{\text{ear}_n}\right)$ where $T_{\text{ear}_n}$ is the estimated availability time of the first available processing unit $U_e$ of the server after the arrival of task $n$ and $Max(.)$ ensures that the processing of the task starts when a free processing unit is found after the offloading of the task is finished. The $T_{\text{ear}_n}$ is calculated based on the completion time of other accepted tasks on the server with the earliest offloading time than that of task $n$. This estimate is adapted from the work of \cite{xiong2023multi}. 

\subsection{Energy Harvesting}
The energy harvesting process is adapted from the work in \cite{9110595}. For simplicity, we assume that the UDs harvest $e_n$ energy at the beginning of each time interval. Initially, each UD is full with its maximum battery capacity of $b_n^{max}$. The level of the battery in the next time interval depends on both energy consumption and harvesting, which evolves according to the following equation in the $T$ time steps. 

\begin{equation}
\label{Battery}
    b_{n}\text{(t+1)} = \min\left(\max\left(b_n\text{(t)} -  E_n\text{(t)} \text{ + } e_n\text{(t)}, 0\right), b_{n}^{\text{max}}\right)
\end{equation}

where $E_n$ is the energy consumption calculated based on $E_{\text{loc}_n}$ and $E_{\text{off}_n}$ as described in Section~\ref{CC:formulation} and $\min\left(.\right)$ and $\max\left(.\right)$ ensure that the level of the battery cannot be negative and does not exceed the maximum capacity. 

\section{Problem Formulation}
\label{CC:formulation}
The total processing latency is $T_n = \left(1 - x_n\right) T_{\text{loc}_n} \text{ + } x_n T_{\text{MEC}_n}$ and the energy consumption is equal to $E_n = \left(1 - x_n\right) E_{\text{loc}_n} \text{ + } x_n E_{\text{off}_n}$. Considering that the cost of processing a task is collectively determined by its processing latency and energy consumption, the cost function for processing a task is specified as $L_n = \lambda_1 T_n \text{ + } \lambda_2  E_n$ where $\lambda_1$ and $\lambda_2$ are weight coefficients. The CCM\_MADRL\_MEC aims to solve the optimization problem that can be formulated as the cost minimization for all UDs and $T$ time steps while meeting the different constraints at the UDs and the server as follows:

\begin{subequations}
\begin{align}
    \underset{\{x_n, p_n, f_n\}}{\text{minimize}} \quad & \bigsum_{t}^{T} \bigsum_{n \in N} L_n\text{(t)} \label{cc:objective} \\
    \text{subject to} \quad & x_n \in \{0, 1\}, \quad \forall n \in N \label{cc:constraint_a} \\
    & p_n^{\min} \leq p_{n} \leq p_n^{\max}, \quad \forall n \in N \label{cc:constraint_b} \\
    & T_n \leq \tau_n, \quad \forall n \in N \label{cc:constraint_c} \\
    & b_n \ge b_n^{min}, \quad \forall n \in N \label{cc:constraint_d} \\
    & f_n^{\min} \leq f_{n} \leq f_n^{\max}, \quad \forall n \in N \label{cc:constraint_e} \\
    & \bigsum_{n \in N} x_n \leq K \label{cc:constraint_f} \\
    & \bigsum_{n \in N} x_n z_n \leq z_e \label{cc:constraint_g}
\end{align}
\end{subequations} where Equation~(\ref{cc:constraint_a}) implies that a task is processed locally or uploaded to the MEC server, Equation~(\ref{cc:constraint_b}) indicates that the transmission power should be between the power allocation budget, Equation~(\ref{cc:constraint_c}) ensures that the processing time of each task cannot exceed its processing deadline, Equation~(\ref{cc:constraint_d}) guarantees that the battery level should not exceed the low battery level, Equation~(\ref{cc:constraint_e}) ensures that the local computational resource allocated to each task should be in the preset minimum and maximum values, Equation~(\ref{cc:constraint_f}) and ensures that the number of offloaded tasks does not exceed the number of sub-channels by ensuring that only one task uses a channel. It is used if and only if it is necessary to use only one channel for one user as used by \cite{8372737}. Equation~(\ref{cc:constraint_g}) guarantees that the sum of the sizes of the off-loaded tasks does not exceed the storage capacity of the server.

\section{Combinatorial Client-Master MADRL Algorithm for Task Offloading in MEC}
\label{CombinatorialClient-Master}
To solve the optimization problem of the cost minimization in Equation~(\ref{cc:objective}), we convert the optimization problem into a reward maximization problem and apply CCM\_MADRL\_MEC. The states, client and master actions, and the formulation of the reward function are presented as follows.

\subsection{State}
The state $S\left(t\right)$ of the MEC environment at time $t$, which includes the set of states of the UDs, is described as $S\left(t\right) = \{S_n\left(t\right)\}, \quad \forall n \in N$. Constant values such as the number of sub-channels $K$, the number of processing units on the server $U_e$, the processing capacity $f_e$, and the storage capacity $z_e$  of the server are excluded from the state information. The state of a UD, $S_n\left(t\right)$, is characterized by five components: task state $S_n^{\text{task}}\left(t\right)$, normalized channel gain state $S_n^{\text{gain}}\left(t\right)$, power transmission budget $S_n^{\text{pow}}\left(t\right)$, local resource allocation budget $S_n^{\text{res}}\left(t\right)$, and battery state $S_n^{\text{battery}}\left(t\right)$ as defined in Equation~(\ref{all_state}):

\begin{equation}
    \label{all_state}
    S_n\left(t\right) = \{S_n^{\text{task}}\left(t\right), S_n^{\text{gain}}\left(t\right), S_n^{\text{pow}}\left(t\right), S_n^{\text{res}}\left(t\right), S_n^{\text{battery}}\left(t\right)\}
\end{equation} where $S_n^{\text{task}}\left(t\right) = \text{[}z_n\left(t\right), c_n\left(t\right), \tau_n\left(t\right) \text{]}$, $S_n^{\text{gain}}\left(t\right)$ is $g_n$ = $h_n$/{$\sigma^2$} as described in Section~\ref{transmitioncapacity}, $S_n^{\text{pow}}\left(t\right)$  = $f_{n}^{max}$, $S_n^{\text{res}}\left(t\right) = f_{n}^{max}$, and $S_n^{\text{battery}}\left(t\right)$ is as described in Equation~(\ref{Battery}).

\subsection{Action}
\label{Bothactions}

At the beginning of each time step, the UDs make decisions about their resource allocations using client agents. Then, the SDN controller collects information about the state and action of the UDs and performs one of the following three procedures using a master agent: 1) for the UDs that decide to make a local processing, the server does not interfere. 2) if the number of UDs that are proposed to offload is greater than the number of sub-channels or if the sum of their sizes is greater than the capacity of the storage capacity of the server, the server makes a combinatorial decision on which of the requests of the UDs to approve and which of them to reject. 3) If the proposed requests are less than the constraints, the server accepts all of them. Finally, sub-channels are assigned to accepted UDs, and then the task offloading and processing process starts. 

Existing DRL-based task offloading algorithms, such as \cite{9573404} and \cite{9829326}, included the number of sub-channels in their state and action space. However, the sub-channels have equal transmission capacity from the perspective of a UD as seen in Equation~(\ref{transmitioncapacity}). If we restrict that a channel is used by only one UD at a time, it does not matter which channel a UD uses. Therefore, the inclusion of sub-channels in the state and action space incurs a dimensionality problem without playing any significant role. We excluded channel information from the state and action space and considered them as a constraint in the combinatorial action selection. Note that a channel can be reused by multiple UDs one after the other, by lifting the constraint that one channel must be used by one channel. In such a case, only the storage capacity of the server becomes the constraint in the combinatorial action selection. 

The actions of the client agents and master agent are as follows. 

\paragraph{Client Actions}At each time step, each client agent produces three actions, which are all continuous value actions between [0, 1] inclusive. The action space can be expressed as:
\begin{equation}
    A_c\left(t\right) = x_{c,n}\left(t\right), p_{c,n}\left(t\right), f_{c,n}\left(t\right) \gets \theta_n\left(S_n\left(t\right)\right), \quad \forall n \in N 
\end{equation} where $S_n\left(t\right)$ is the state of UD $n$ as described in Equation~(\ref{all_state}), and $\theta_n$ is the parametrized policy function of the client, $x_{c,n}\left(t\right)$ is the task offloading decision by client $n$ (if $x_{c,n}\left(t\right) < 0.5$ then $x_n$ in Equation~(\ref{cc:x_n}) becomes 0 otherwise the task is proposed to be considered for the combinatorial decision), $p_{c,n}\left(t\right)$ is the client action that decides the transmission power using Equation~(\ref{decidep}), and $f_{c,n}\left(t\right)$ is the action that decides the local computational resources allocation using Equation~(\ref{decidef}).

The actions of the client agents determine $p_{n}$ and $f_{n}$ are follows:
\begin{equation}
    \label{decidep}
    p_{n} = \max\left(p_{n}^{min},p_{c,n}\left(t\right)p_{n}^{max}\right)
\end{equation}
\begin{equation}
    \label{decidef}
    f_{n} = \max\left(f_{n}^{min},f_{c,n}\left(t\right)f_{n}^{max}\right)
\end{equation}

\paragraph{Master Action}For the client actions with $x_{c,n}\left(t\right) \ge 0.5$, the master takes the combinations of states and actions of the clients and provides a binary output for the combinatorial decision on which of them should be allocated locally and which of them should be accepted for processing by the MEC server. 

\begin{equation}
    A_m\left(t\right) = x_{m,n}\left(t\right) \gets \phi\left(S, A, S_c, A_c\right)
\end{equation} where $S$ and $A$ are the set of states and actions of all client agents, and $S_c$ and $A_c$ are the set of states and actions of the client agents whose $A_{c,n} \geq 0.5$ and $\phi$ is the policy of the master. 

The combinatorial action selection in the master agent is built by modifying the critic in the classical multiagent deep deterministic policy gradient (MADDPG) algorithm~\cite{lowe2017multi} and incorporating the per-action DQN~\cite{He2016}. 

\begin{figure}[t]
  \centerline{\includegraphics[width=1\linewidth]{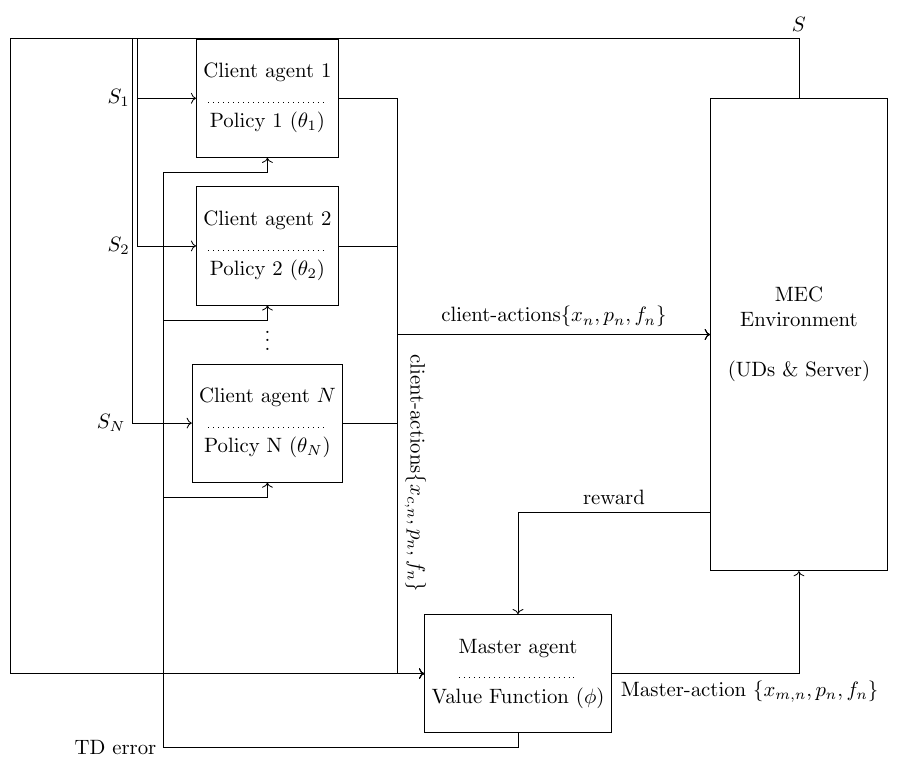}}
  \caption{The interaction diagram of the agents and the MEC environment. Client agents output their actions $\{x_{n}, p_n, f_{n}\}$. Clients with $\{ x_{c,n} <  0.5$\} start local processing; and the others propose their tasks to the master agent, who makes the combinatorial decision on which of the proposed tasks should be offloaded and which of them should be designated for local processing}
  \label{fig:CCM_MADRL}
\end{figure}

Figure~\ref{fig:CCM_MADRL} shows the interaction diagram of the CCM\_MADRL algorithm and the MEC system. Client agents represent the policies of the UDs. The master agent represents the policy on the MEC server. The environment represents the allocation of resources in the UDs and on the server. After a client produces its output, it does the following as mentioned in Section~\ref{Bothactions}: if $x_{c,n} < 0 $ assigns $x_{n} = 0$ and starts the local allocation. Otherwise, it forwards $ x_{c,n}, p_n, f_{n}$ to the master for the combinatorial decision. Then, the master agent produces the binary decision and applies it to the UDs and the server. Finally, a shared reward is computed and provided to the master agent to train its value function. The client agents are also trained using a TD error computed by the master agent as feedback.

\subsection{System Reward Function}
To compute the reward, we use the negative of the objective function and we compute a penalty function for the constraints of the tasks and UDs. The constraints of the server are already considered in the combinatorial action selection and do not need to be incorporated as a penalty. In \cite{9110595}, they included a dropoff penalty in their reward for running out of batteries. In ours, we start the penalty from a preset minimum battery threshold. 

\begin{equation}
    L'_n = \lambda_1 \min\left(\left(\tau_n - T_n\right),0\right) \text{ + } \lambda_2  \min\left(\left(b_n - b_n^{min}\right),0\right)
\end{equation}

Since the design of our cooperative learning formulation is based on the cost minimization problem in Equation~(\ref{cc:objective}), our system reward function is equal to the negative of the system cost function and the penalty function. Thus, we can formulate the system reward function as follows:

\begin{equation}
\bar{r}\left(S\left(t\right), A\left(t\right)\right) = - \frac{1}{|N|} \bigsum_{n \in N} (L_n\left(t\right)-L'_n\left(t\right)
\label{comreward}
\end{equation}

The power and resource allocation decisions made in the current step by the UDs affect their operational life in the next time steps by affecting energy consumption. Therefore, the DRL must consider immediate reward and long-term return using the Bellman equation as shown in Equation~(\ref{CCM_MADRL_MEC:Bellman})
\begin{equation}
    \begin{aligned}
    Q\left(S, A, S_n, A_n|\phi\right) &= \left(1-\alpha_\phi\right)Q\left(S, A\right) \text{ + } \alpha_\phi \left( R\left(S, A\right) \right. \\
    & \quad \text{ + } \gamma \sum_{n} P\left(S' | S, A\right) \max_{S'_n, A'_n} Q\left(S', A', S'_n, A'_n|\phi'\right)\left.\right)
    \end{aligned}
    \label{CCM_MADRL_MEC:Bellman}
\end{equation}

where $\alpha_\phi$ is the learning rate, $\gamma$ is the discount factor, $\phi$ is the policy of the critic, $\phi'$ is the policy of the target critic, $S$ = \{$ S_1,..., S_N$\} and $S'$ = \{$ S'_1,..., S'_N$\} are combined current and next states, $A$ = \{$ A_1,..., A_N$\} and $A'$ = \{$ A_1,..., A'_N$\} are combined current and next actions of the client agents, and $S'_n$ and $A'_n$ are the corresponding states and actions of the agents.  The role of the four parameters in $Q'$ is presented in the following section. 

\subsection{The master Agent with per-client DQN}
\label{TheMaster}
In MADDPG, there is only a single Q-value for the combined state and action pair of all actors, which is calculated as:
\begin{align}
 Q\left(S, A\right) = \left(1-\alpha_\phi\right)Q\left(S, A|\phi\right) \text{ + } \alpha_\phi \left( R\left(S, A\right) \text{ + } \gamma Q\left(S', A'|\phi'\right)\right)
\end{align}

To customize the critic to select combinatorial actions, it should be able to provide a Q-value per each actor (or client in the case of CCM\_MADRL). Therefore, the master agent with per-client DQN in the CCM\_MADRL adapts the concept of per-action DQN where the state $S_n$ and the action $A_n$ of each client agent are appended to the combined state and the action of the client agents to calculate the relative Q value in the combination of the state and the action as seen in Equation~(\ref{CCM_MADRL_MEC:Bellman}). The combined rewards are given only to selected clients in the task offloading problem, as seen in Algorithm~\ref{alg:TraingAlgorithm} so that they will have different Q values to distinguish them in action selection. The reward is used by the master agent to train a value function for the selected clients using per-client DQN. Client agents are also trained using a TD error computed by the master agent as feedback.

The master agent applies the coalition action selection approach in \cite{10.5555/3635637.3662918}. However, for ease of benchmark comparison with MADDPG, it uses per-action DQN instead of a transformer neural network.

\subsection{Algorithms}

Since the master agent in the CCM\_MADRL\_MEC algorithm has two roles: providing feedback for training the clients, similar to the MADDPG, and participating in the combinatorial action selection of the clients, it follows different procedures for both. Therefore, the algorithm is presented in three parts: a main algorithm, an action selection algorithm, and a training algorithm in the following sections. 

\paragraph{Main Algorithm}
The main algorithm runs the action selection algorithm, the training environment, and the evaluation environment. At each episode, it runs for $T$ time steps as seen in lines 7 to 12 of Algorithm~\ref{alg:CCMMADRLAlgorithm}. At each step, the action selection algorithm is called, and the rewards are computed, and then the experience is recorded to replay memory. When the iteration over the steps is complete, the training algorithm is called and the trained policies of the client agents and the master agent are evaluated.

\begin{algorithm}
\caption{CCM\_MADRL main algorithm}
\label{alg:CCMMADRLAlgorithm}
\begin{algorithmic}[1]
\STATE Initialize $Max\_Episodes$ = 2000, $Min\_Epsilon$ = 0.01, $Max\_Epsilon$ = 1, $\gamma$ = 0.99
\STATE Initialize client agents $\theta_n \quad \forall n \in N$ and the master agent $\phi$ with random weights
\STATE Initialize target client agents $\theta'_n \gets \theta_n$ and the target master agent $\phi' \gets \phi$, $\quad \forall n \in N$
\STATE Initialize replay memory $RM$
\FOR{$\text{episode} = 1$ to $Max\_Episodes$}
    \STATE Reset environment and get initial state $S_n\left(t=1\right)$, $\quad \forall n \in N$
    \FOR{$t = 1$ to $T$}
        \STATE Go to Algorithm~\ref{alg:ActionSelectionAlgorithm} using evaluation = False flag to select client and master actions
        \STATE Execute actions and observe total reward $\bar{r}\left(t\right)$ and next state $S_n\left(t \text{ + }1\right)$, $\quad \forall n \in N$
        \STATE Store transition $\left(S_n\left(t\right), A_n\left(t\right), \bar{r}\left(t\right), S_n\left(t \text{ + } 1\right)\right)$, $\quad \forall n \in N$  in to $RM$
        \STATE Update the state $S_n\left(t\right) \gets S_n\left(t \text{ + }1\right)$, $\quad \forall n \in N$
    \ENDFOR
    \STATE Go to Algorithm~\ref{alg:TraingAlgorithm} for training
    \FOR{$\text{EvalEpisode}$ in $\text{EvalEpisodes}$}
        \STATE Reset and seed episode to $\text{EvalEpisode}$ and find state $S\left(t=1\right)$
        \FOR{$t = 1$ to $T$}
            \STATE Go to Algorithm~\ref{alg:ActionSelectionAlgorithm} using evaluation = True flag to select client and master actions
            \STATE Execute actions and observe total reward $\bar{r}\left(t\right)$ and next state $S_n\left(t \text{ + }1\right)$, $\quad \forall n \in N$
            \STATE Update the state $S_n\left(t\right) \gets S_n\left(t \text{ + }1\right)$ , $\quad \forall n \in N$
        \ENDFOR
    \ENDFOR
\ENDFOR
\end{algorithmic}
\end{algorithm}

\paragraph{Action Selection Algorithm}

Because the action selection algorithm has to use exploration in the training environment, we present the computation of $\epsilon$, which is used for $\epsilon$-greedy to determine whether to explore a new action or to exploit the learned knowledge in the master agent and to scale the noise in the client agents as seen in Algorithm~\ref{alg:ActionSelectionAlgorithm}. At each episode, $\epsilon$ is updated using Equation~(\ref{ccm_madrl_mec_epsilon}).

\begin{equation}
    \epsilon = Min\_Epsilon \text{ + } \left(Max\_Epsilon - Min\_Epsilon\right) \cdot e^{-\frac{episode}{Max\_Episodes}}
    \label{ccm_madrl_mec_epsilon}
\end{equation} where $Min\_Epsilon$ and $Max\_Epsilon$ are the minimum and maximum values of the decaying epsilon, $episode$, is the current episode, and $Max\_Episodes$ is the maximum number of episodes.

The action selection algorithm applies the exploration of actions for the client agents and the master agent as follows. Note that the evaluation flag is used to indicate whether the actions are running for the training environment or for the evaluation environment. In the evaluation, exploration is not needed. For client agents, the exploration is performed by adding noise to the actual output of the client agents, as seen in lines 5 to 8 of the Algorithm~\ref{alg:ActionSelectionAlgorithm}. After adding noise to the actual action, the values are clipped to [-1,1] so that they are within the activation function of the client agents, Tanh in this case. All actions, explored or exploited, are scaled to be between [0,1] before applying to compute the resource allocation in Section~\ref{Bothactions}. 

The master agent follows $\epsilon$-greedy for exploration and exploitation as seen in lines 41 to 47. First, a random number is generated as seen in line 11 to decide whether to explore or exploit. If the number is less than $\epsilon$, the master agent shuffles the proposed actions as seen in line 45 and follows the combinatorial action selection procedure described below. Otherwise, the master agent computes the Q value based on the states and actions of the proposed actions and appends the Q value along with the identifiers of the tasks $n$ to $Qs$ and $Index$ and follows the combinatorial action selection procedure. 

After the actions of the client agents are provided, the master agent follows one of the following three procedures in the exploitation mode as described in Section~\ref{Bothactions}. If all client agents decide to process their tasks locally, as in line 18, the master agent does not intervene. Line 20 computes the Q values of the proposed tasks using per-client DQN, and appends them to $Qs$ and $Index$ along with their identifiers $n$, to be considered in the combinatorial decision. If the number of proposed tasks or the sum of their sizes is less than the number of sub-channels and the storage constraint on the server, the server accepts all of them as seen in lines 24 to 26. If the number of proposed tasks is greater than the number of sub-channels or if the sum of their size is greater than the storage capacity of the server, the master agent uses the Q-values computed using the states and actions of the client agents that proposed to offload their tasks to make decisions. It starts to approve the proposed actions of the clients with the highest Q-values until the number of sub-channels or the storage constraint is met. The remaining agents are designated to process their tasks locally. The algorithm is provided in Algorithm~\ref{alg:ActionSelectionAlgorithm}. The procedure of the exploitation is provided similarly to the procedure of exploitation except that the proposed actions are shuffled randomly rather than getting sorted by their Q-values.
\begin{algorithm}
    \caption{The action selection algorithm for the client agents and the master agent}
    \label{alg:ActionSelectionAlgorithm}
    \begin{algorithmic}[1]
        \STATE Input: state $S_n$ for each client agent $n$, $\{K,z_e\}$, and Evaluation flag
        \STATE Output: client actions $A_n$ for each $n$ with $x_n$ decided by collaboration with the master agent
        \STATE Get action $A_n \gets \pi_n\left(S_n, \theta_n\right)$, $\quad \forall n \in N$
        \IF{Evaluation == False}
            \STATE Compute $\epsilon$ using Equation~(\ref{ccm_madrl_mec_epsilon})
            \STATE $noise$ = random($|N|$ by $|A_n|$)*$\epsilon$
            \STATE $A_n$ = $A_n$ + $noise_n$, $\quad \forall n \in N$
            \STATE Clip $A_n$ to [-1, 1]            
        \ENDIF
        \STATE Scale $A_n$ to [0, 1] using $\frac{A_n}{2} \text{ + } 0.5$, $\quad \forall n \in N$
        \STATE Generate a $random$ number
        \STATE $Qs$ = [], $Index$ = []
        \IF{$random < \epsilon$ or Evaluation == True}
            \STATE $S = \{S_n\}$, $A = \{A_n\}$, $\quad \forall n \in N$
            \FOR{$ n \in N$}
                \STATE Get $x_{c,n}$ from $A_n$ as described in Section~\ref{Bothactions}
                \IF{$x_{c,n} < 0.5$}
                    \STATE $x_n$ = 0
                \ELSE
                    \STATE $Append\left(Qs,Q\left(S,A,S_n,A_n, \phi\right)\right)$, $Append\left(Index,n\right)$
                \ENDIF
            \ENDFOR
            \IF {$Length\left(Index\right) \leq K $ and $Sum\left(z_n \; \forall n \in N \; and \; x_{c,n} \ge 0.5\right)$ $ \leq z_e$ }
                \FOR{$\forall n \in N \; and \; x_{c,n} \ge 0.5$}
                    \STATE $x_n$ = 1 
                \ENDFOR
            \ELSE
                \STATE Sort $Qs$, and adjust $Index$ accordingly
                \STATE $TotalSizeOfAccepted $ = 0
                \WHILE{$Length\left(Index\right) > K$}
                    \STATE $n$ = $Pop\left(Index\right)$
                    \STATE $x_n$ = 1 
                    \IF{$TotalSizeOfAccepted $ + $z_n$ $ \leq z_e$}
                        \STATE $TotalSizeOfAccepted$ = $TotalSizeOfAccepted$ + $z_n$
                    \ELSE
                        \STATE $x_n$ = 0
                    \ENDIF
                \ENDWHILE
            \ENDIF
        \ELSE
            \STATE Collect the index $n$ of the tasks with $x_{c,n} \ge 0.5$ $\forall n \in N$ to $Index$
            \IF {$Length\left(Index\right) \leq K $ and $Sum\left(z_n \; \forall n \in N \; and \; x_{c,n} \ge 0.5\right)$ $ \leq z_e$ }
                \STATE Execute lines 24 to 26
            \ELSE
                \STATE shuffle $Index$ in to random order
                \STATE Execute lines 29 to 38
            \ENDIF
        \ENDIF
    \end{algorithmic}
\end{algorithm}

\paragraph{Training Algorithm}

\begin{algorithm}
\caption{The training algorithm for the client agents and the master agent}
\label{alg:TraingAlgorithm}
\begin{algorithmic}[1]
\STATE Sample a random minibatch of transitions $\left(S, A, A^{mas}, \Bar{r}, S', done\right)$ of size $M$ from $RM$
\STATE Set target actions $A'_{i,n} \gets \pi_n\left(S'_{i,n}, \theta'_n\right)$, $\quad \forall n \in N$, and for $i = 1$ to $M$
\STATE $S'_i = \{S'_{i,n}\}$, $\quad \forall n \in N$, and for $i = 1$ to $M$
\STATE $A'_i = \{A'_{i,n}\}$, $\quad \forall n \in N$, and for $i = 1$ to $M$
\STATE Get $x_{i,n}$ from $A'_{i,n}$ as described in Section~\ref{Bothactions}, $\quad \forall n \in N$, and for $i = 1$ to $M$
\IF{$x'_{i,n} \ge 0.5$,$ \quad \forall x'_{i,n} \in a'_{i,n}$, $\quad \forall n \in N$, and for $i = 1$ to $M$}
    \STATE $Append\left(Q'_N, Q\left(S'_i,A'_i, S'_{i,n}, A'_{i,n}, \phi'\right)\right)$
\ENDIF
\STATE $Q'_i = Q'_N$, for $i = 1$ to $M$
\IF{$Length\left(Q'_i\right)$ is 0 for any $i$ }
    \STATE $nextQ_i$ = $Q\left(S'_i,A'_i, all\_zeros, all\_zeros, \phi'\right)$
\ELSE
    \STATE $nextQ_i = Max\left(Q'_i\right)$
\ENDIF
\STATE y = [], Qs = []
\IF{$A_{i,n}^{mas} = 1$ $\quad \exists n \in N \quad \exists i \in M$}
    \STATE $Append\left(Qs, Q\left(S_i,A_i, S_{i,n}, A_{i,n}, \phi\right)\right)$ 
    \STATE targetQ = $ \Bar{r_i} \text{ + } \gamma nextQ_i*\left(1-done_i\right)$ 
    \STATE append(y, targetQ)
\ENDIF
\IF{$A_{i,n}^{mas} = 0$ $, \quad \forall n \in N \quad \exists i \in M$ }
        \STATE $Append\left(Qs, Q\left(S_i,A_i, all\_zeros, all\_zeros, \phi\right)\right)$ 
        \STATE targetQ = $ \Bar{r_i} \text{ + } \gamma nextQ_i*\left(1-done_i\right)$ 
        \STATE $append\left(y, targetQ\right)$
\ENDIF
\STATE Compute the TD error: $\delta = \frac{1}{Length\left(y\right)} \bigsum_{j=1}^{Length\left(y\right)} \left(y_j - Qs_j\right)^2$
\STATE Update parameters of master agent $\phi$: $\phi \gets \phi \text{ + } \alpha_{\phi} \cdot \nabla_\phi \delta$
\STATE Update target master network $\phi' \gets \phi $       
\FOR{each client $n$}
    \STATE $Q^N_i$ = [] for $i = 1$ to $M$, $tarQ$ = []
    \STATE Set new actions $A^{new}_{i,n} \gets \pi_n\left(S_{i,n}, \theta_n\right)$, $\quad \forall n \in N$, and for $i = 1$ to $M$
    \STATE $A^{new}_i = \{A^{new}_{i,n}\}$, $\quad \forall n \in N$, and for $i = 1$ to $M$
    \STATE $Append\left(Q^N_i, Q\left(S_i,A^{new}_{i}, S_{i,n}, A^{new}_{i,n}, \phi\right)\right)$, $\quad \forall n \in N$ with $A^{new}_{i,n} \ge 0$, and for $i = 1$ to $M$
    \IF{$Length\left(Q^N_i\right)$ is 0 for any $i$}
        \STATE $Qloc$ = $Q\left(S_i, A^{new}_{i}, all\_zeros, all\_zeros), \phi\right)$
        \STATE $Append\left(tarQ, Qloc\right)$
    \ELSE
        \STATE $Append\left(tarQ, Max\left(Q^N_i\right)\right)$)
    \ENDIF
    \STATE Compute the gradient for the client: $\nabla_{\phi_n} J(\phi_n) \gets -\frac{1}{Length(tarQ)} \bigsum_{j=1}^{Length\left(tarQ\right)} \nabla_{\phi_n}tarQ_j$
    \STATE Update the client parameters $\theta_n$:
    \STATE \hspace{2em} $\theta_n \gets \theta_n - \alpha_{\theta} \nabla_{\theta_n} J\left(\theta_n\right)$
    \STATE Update target client networks $\theta'_n \gets \theta_n$
\ENDFOR
\end{algorithmic}
\end{algorithm}
The algorithm for training the client agents and the master agent is provided in Algorithm~\ref{alg:TraingAlgorithm}. Because the structure of the master agent is different from the MADDPG critic as seen in Section~\ref{TheMaster}, Algorithm~\ref{alg:TraingAlgorithm} is significantly different from existing MADDPG training algorithms in that: It generates multiple Q-values rather than one combined Q-value, because the master agent has to make a combinatorial decision using the relative Q-values of the clients as seen in lines 7,17, and 33; The client agents are trained by computing the highest Q value from the tasks offloaded to the server as seen in lines 13 and 38;  If all tasks are allocated locally by the client agents, the master agent uses $all\_zeros$ as a placeholder to hold the combined Q-value as seen in lines 11, 22, and 35. Note that after the master agent decides which of the tasks should be processed in the server as described in Section~\ref{Bothactions}, the reward is applied only to the selected tasks during training. That is, the reward is computed at a system level using Equation~(\ref{comreward}) but it is used only by the tasks offloaded to the server when training the algorithm so that the tasks are distinguished by their Q-values in the action selection algorithm. This technique is adapted from the coalition action selection in \cite{10.5555/3635637.3662918}. 

The notation of the client and master actions is changed in the training algorithm due to the subscript $i$ for the minibatch which is used to iterate over the entries of the minibatch of size $M$. Unlike MADDPG, which computes the Q-values of the minibatch as a batch, the Q-values of the minibatch in the CCM\_MADRL are computed individually because they are processed conditionally as seen with many if clauses in the algorithm. The training algorithm starts by selecting a minibatch of size $M$ from the replay memory. Each entry in the minibatch includes the combined state $S$ and action $A$ of all client agents $S$, the set of binary actions of the tasks by the master agent$ A^{mas}$, the total reward of the tasks $\Bar{r}$, the combined next state $S'$, and a flag that indicates whether the episode was terminated or not $done$.

The master agent is trained by lines 2 through 28. Line 2 computes the target action for every client and every entry of the minibatch using their next state. The target action is to be used to compute the target Q-value using the master agent. Then, lines 3 and 4 concatenate the target actions of the client agents because the master agent accepts a combined state and action of all clients as input as seen in line 7. Lines 5 and 6 check if the client agents have decided to process the tasks locally or propose them to the master agent. For each task that was proposed to the master, a relative Q-value is computed on line 7 and the maximum Q-value will be computed in line 13. Line 9 concatinates the Q-values of the offloaded tasks in the same entry of a minibatch. If no task was offloaded, a Q-value will be computed using a placeholder to train the master agent so that it is used to give feedback like the classical MADDPG. Then, the combined reward is provided to the actions selected to offload their tasks as seen in lines 16 to 20. The current and target Q-values in lines 17 and 17 are used to compute the TD error in line 26. As seen in lines 21 to 25, if all client agents, at any entry of the minibatch, decide to process their tasks locally, the master agent concatenates the state and action of all agents and appends $all\_zeros$ as a placeholder to learn the Q value when all tasks are processed locally, and it is only used to provide feedback in training the client agents as seen in line 35.   

The training of client agents is seen from lines 29 to 44. They are trained similarly to the training of actors in classical MADDPG except that the feedback is computed differently as seen in lines 33 to 39, because the Q value is provided for the client agents that offloaded their task to the server. Therefore, if one or more clients were offloaded their task, the feedback for training the clients is computed from the Q value of one of the offloaded tasks as they are trained with the same rewards. The maximum Q value of the offloaded tasks is considered for consistency. The calculation of the maximum Q-value is the same as that of the training for the master agent.

We used DDQN~\cite{van2016deep} and prioritized experience replay~\cite{schaul2015prioritized} in the CCM\_MADRL algorithms for better efficiency in the training. 

\section{Experimental Evaluation}
\label{CC:experimet}
To evaluate the merits of the combinatorial action selection by introducing a master agent to MADDPG, in a task offloading problem with various constraints, we compare our algorithm with other benchmarks and heuristic algorithms as follows.  

\subsection{Benchmark Comparison}
 The main benchmark for our algorithm is MADDPG for two reasons. First, most existing task-offloading algorithms use DDPG and MADDPG. Second, our algorithm is an extension of MADDPG. However, we also developed different heuristic benchmarks. The heuristics differ from the proposed CCM\_MADRL\_MEC in that, instead of training a master agent to make combinatorial decisions about the clients, a stationary algorithm is used to decide on which of the clients to approve for the MEC server based on some ordering mechanism. The benchmark algorithms are discussed below.  
\begin{itemize}
    \item MADDPG: This benchmark uses actor agents to make decisions. Its difference from CCM\_MADRL\_MEC is in the procedure 2) of the action selection in Section~\ref{Bothactions}, where the SDN allocates the tasks to the sub-channels according to their order of offloading rather than making combinatorial decisions. Tasks that are not assigned to any channel are dropped. The actor agent in the UD will assign $\tau_{max}$ to its $T_{\text{MEC}_n}$ for the dropped tasks as a penalty. Because the penalty can be unfair for benchmark comparison, the following heuristics are developed to have equivalent combinatorial decisions with the CCM\_MADRL\_MEC for the tasks that are not accepted. 
    \item MADDPG with the shortest offloading time first heuristic: This is similar to MADDPG, with the distinction that tasks not assigned to sub-channels or storage are designated for local processing. 
    \item MADDPG with deadline/size first heuristic: This differs from the MADDPG with the shortest offloading time first heuristic in that it uses the increasing order of deadline/size as a priority rather than offloading time.
\end{itemize}

Even if the way tasks are accepted by the MEC server differs between the benchmarks and the CCM\_MADRL\_MEC, the order of processing of the accepted tasks is always in the order of arrival at the MEC using Equation~(\ref{Tappo})

\subsection{Experimental Settings}

The experimental setting is provided in Table~\ref{Taskoffsetting}. Note that all UDs have the same minimum battery, power, and resource allocation threshold, but their maximum budget is generated from a uniform distribution. As described in Section~\ref{CC:sytemmodle}, the experimental setting is customized from the settings in \cite{9573404} and \cite{9110595}. The typical storage capacity of modern servers is GB and TB. However, because we chose a small experimental setting due to computational resources, we considered a storage constraint of 400 MB so that the task offloading problem is combinatorial to the server. Evaluation episodes are seeded with their index. We used a seed of 37 for the reproducibility of the simulation environment. The source code is publicly available at \href{https://github.com/TesfayZ/CCM_MADRL_MEC}{\textcolor{blue}{https://github.com/TesfayZ/CCM\_MADRL\_MEC}}. 

The hyperparameters for the CMMADRL and benchmark algorithms are as follows: a discount factor of 0.99, a minibatch size of 64, a replay memory size of 10,000, and learning rates of 0.0001 for the clients and actors and 0.001 for the master and critic networks. The output activation function is Tanh for the clients and actors, while ReLU is used for the master and critic networks.

The neural network architecture for the clients and actors consists of an input layer with 7 neurons, two hidden layers with 64 and 32 neurons, and an output layer of size 3. The critic network in the benchmark algorithms has an input layer of 500 neurons, representing the combined state and action size for the 50 actors, followed by two hidden layers with 512 and 128 neurons, and a single output neuron. The master agent has similar parameters to the critic, but its input layer has 510 neurons to account for computing the Q-value for each client using a per-action DQN~\cite{He2016}. All input and hidden layers use linear activation functions. 

\begin{table}[htbp]
\begin{center}
\caption{Experimental parameters of CCM\_MADRL\_MEC}
\label{Taskoffsetting}
\begin{tabular}{|p{0.1\linewidth}|p{0.3\linewidth}|p{0.1\linewidth}|p{0.3\linewidth}|}
\hline
\textbf{Param} & \textbf{Value} & \textbf{Param} & \textbf{Value} \\
\hline
$|N|$ & 50 & $P^{max}$ & 24 dBm \\
$K$ & 10 & $P^{min}$ & 1 dBm \\
$\tau_n$ & [0.1-0.9] s & $p_n^{max}$ & [$P^{min}$ - $P^{max}$] dBm \\
$W$ & 40 MHz & $p_n^{min}$ & $P^{min}$ dBm \\
$f_{max}$ & 1.5 GHz & $f_{min}$ & 0.4 GHz \\
$f_n^{max}$ & [$f^{min}$- $f^{max}$] GHz & $b^{max}$ & 3.2 MJ \\
$f_n^{min}$ & $f^{min}$ GHz& $b^{min}$ & 0.5 MJ \\
$g_{n}$ & [5-14] dB & $f_e$ & 4 GHz \\
$z_n$ & [1-50] MB & $z_e$ & 400 MB \\
$c_n$ & [300 - 737.5] cycles & $U_e$ & 8 \\
$\kappa$ & $5 \times 10^{-27}$ & $\lambda_1, \lambda_2$ & 0.5, 0.5 \\
$e_n$ & 0.001 J & $b_n^{max}$ & [$b^{min}$- $b^{max}$] MJ\\
$b_n^{min}$ & $b^{min}$ MJ& &\\
\hline
\end{tabular}
\end{center}
\end{table}

\subsection{Generalizability}
\label{Generalizability}

Because convergence is affected by the initialization of the weights of the DNNs and the exploration and exploration sequence, we evaluate the algorithms in a different evaluation environment. At each episode of the training environment, the DRL is evaluated with 50 evaluation episodes.

\subsection{Results}

\begin{figure}
\centerline{\includegraphics[width=1\linewidth]{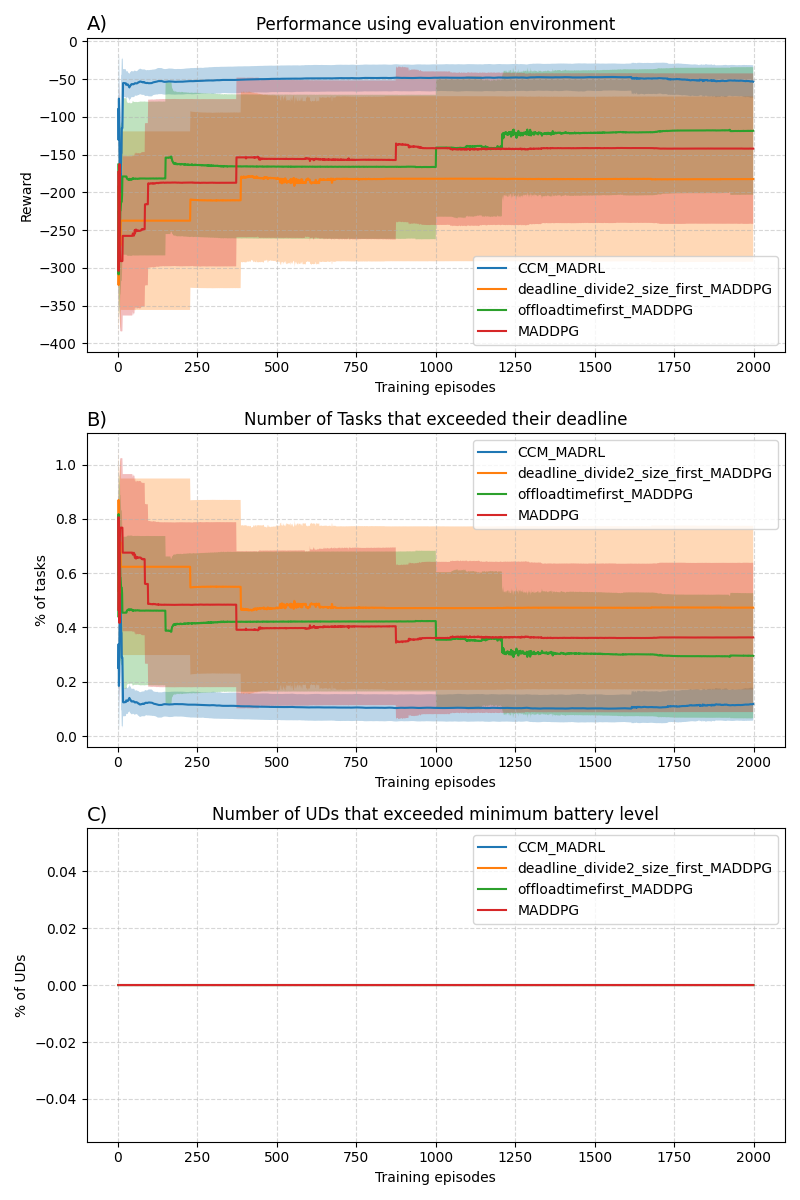}}
\caption{Comparison of the CCM\_MADRL in the evaluation environment with the heuristic and MADDPG algorithms with a learning rate of 0.0001 and 0.001 for the clients and master respectively}
\label{s10Evallre4e3}
\end{figure}

First, we run an experiment with 10 steps per episode for 2000 training episodes. We performed 10 experiments using different initializations of DNN weights and different sequences of exploration and exploitation for each run and plotted the result with a 95\% confidence interval as seen in Figure~\ref{s10Evallre4e3} (A). The CCM\_MADRL algorithm has performed better than the others because once the clients choose their action, the master agent also makes a combinatorial decision on the action of the clients.  On the other hand, MADDPG-based heuristic and benchmark algorithms use only actors to provide actions. The critic is only used to provide feedback to the actors. Similarly, Figure~\ref{s10Evallre4e3} (B) shows  CCM\_MADRL has a smaller percentage of tasks that have expired before completing the processing of the tasks compared to the other algorithms. Note that Figure~\ref{s10Evallre4e3} (A) is the performance based on the combined reward and the combined penalty for the time and energy consumption in Equation~(\ref{comreward}) but Figure~\ref{s10Evallre4e3} (B) is a percentage of tasks whose deadline expired before completing their processing to the total number of tasks generated in the episode. Figure~\ref{s10Evallre4e3} (C) shows that none of the UDs exceeds the minimum battery threshold in an episode. This is because the experiments are run for 10 steps per episode.

\begin{figure}
\centerline{\includegraphics[width=1\linewidth]{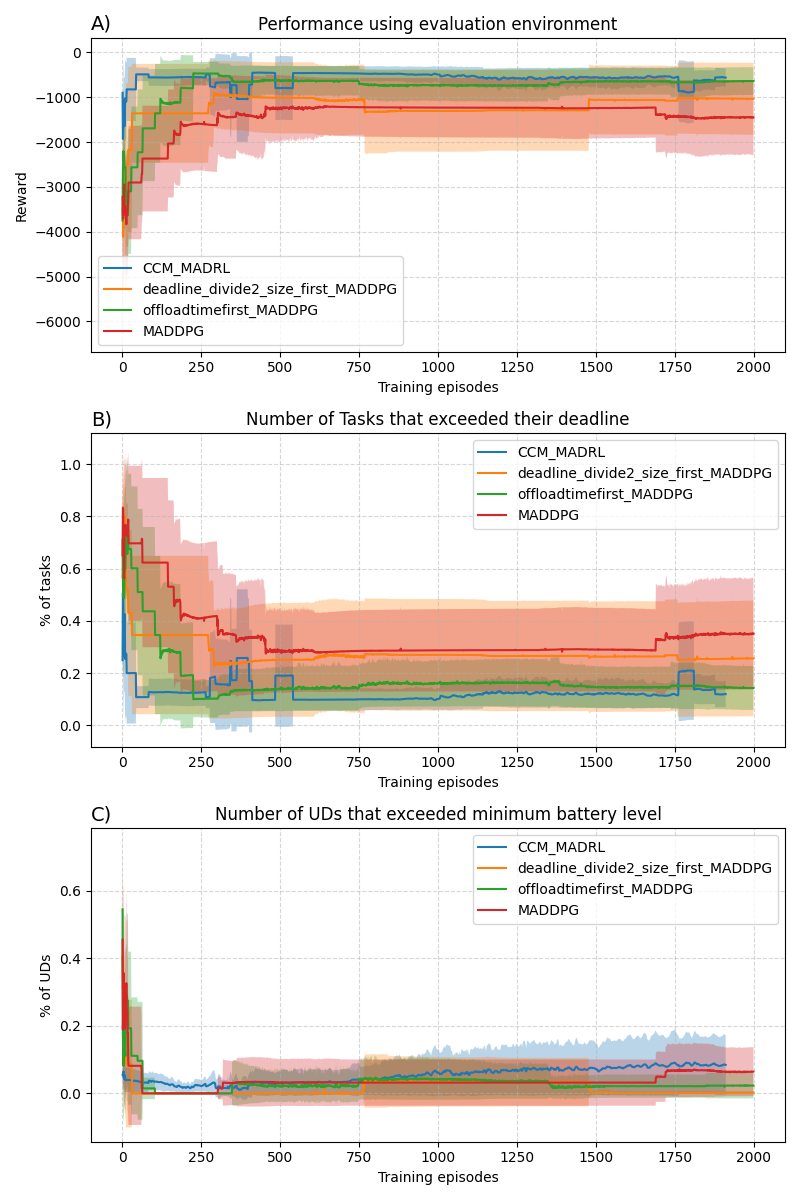}}
\caption{Comparison of the CCM\_MADRL with the heuristic and MADDPG algorithms on the evaluation environment with learning rates of 0.0001 and 0.001 for the clients and master respectively, and $b_{max}$ = $b_{min}$+$1J$}
\label{s100AtEval}
\end{figure}

\begin{figure}
\centerline{\includegraphics[width=1\linewidth]{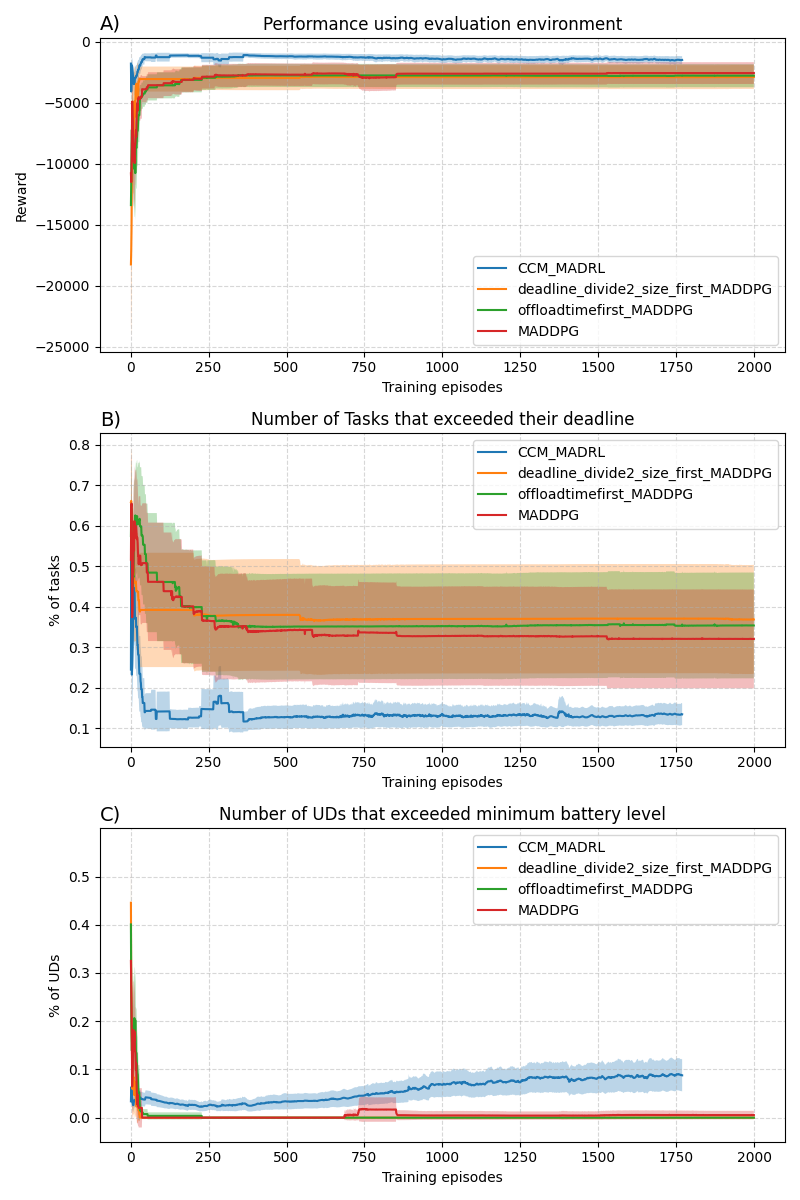}}
\caption{Comparison of the CCM\_MADRL with the heuristic and MADDPG algorithms on the evaluation environment with learning rates of 0.0001 and 0.001 for the clients and master respectively, $b_{max}$ = $b_{min}$+$1J$, and $\lambda_2$ = 5}
\label{s100AtEvalLambda5}
\end{figure}

Next, to see the impact on the battery level, we repeat the above experiment by changing the number of steps per episode to 100 and the $b_{max}$ to $b_{min}$+$1J$. Figure~\ref{s100AtEval} (C) shows that CCM\_MADRL\_MEC has more UDs running below the minimum battery threshold than the benchmark and heuristic algorithms. However, as explained above for Figure~\ref{s10Evallre4e3}, the algorithm is trained by computing a scalar reward, which is a sum of energy and time consumption, as seen in Equation~(\ref{comreward}) but the percentage of UDs that exceed their battery threshold is from the ration of UDs that exceed the battery threshold to the total number of UDs. Therefore, it is affected by the scales given to the deadline penalty and the energy penalty. We used $\lambda_1$ = $\lambda_2$ = 0.5 as weight coefficients for processing time and energy consumption. We repeated the algorithm for $\lambda_1$ = 1 and $\lambda_2$ = 5 as seen in Figure~\ref{s100AtEvalLambda5}.

We further run the experiment with $\lambda_1$ = 1 and $\lambda_2$ = 5 because Figure~\ref{s100AtEval} (B) looks like it is the exact inverse of Figure~\ref{s100AtEval} (B) means $\lambda_1$ = $\lambda_2$ = 0.5 has not given enough balance for the energy consumption, which is a very low number, and the time consumption, which is relatively larger. The result is plotted in Figure~\ref{s100AtEvalLambda5} with a 95\% confidence interval of 40 runs, unlike the above two experiments which are plotted with a 95\% confidence interval of 10 runs. However, the plots do not show a significant difference with 10 runs and 40 runs. Subplot (B) and subplot (A) are now varied, but subplot (C) shows that CCM\_MADRL\_MEC has still more UDs running below the minimum battery threshold than the benchmark and heuristic algorithms. Nonetheless, CCM\_MADRL\_MEC has performed better than the other algorithms in the combined performance and shows more advantage in the number of tasks whose deadlines exceeded. Using scalar reward for multi-objective functions does not represent the underlying problem \cite{vamplew2022scalar}, but it is not within the scope of this work. The work by \citet{hayes2022practical} has studied the essence and techniques of multi-objective reinforcement learning. 

It can be seen that the benchmark and heuristic algorithms demonstrated a closer performance to the CCM\_MADRL\_MEC on 100 steps per episode than on 10 steps per episode. This is because training is performed at the end of each episode. This caused the training after 10 steps to overfit to the early episodes, whereas the training after every 100 episodes to generalize. The benchmark and heuristic algorithms are impacted by overfitting more than CCM\_MADRL\_MEC because they use only their actors to select action, while the CCM\_MADRL\_MEC uses the advantage of both clients and master to mitigate overfitting and sticking to local optimal. 

Note that the CCM\_MADRL\_MEC in Figures~\ref{s100AtEval} and ~\ref{s100AtEvalLambda5} is not plotted until the last episode. The experiment is run on Iridis~\footnote{\href{https://www.southampton.ac.uk/isolutions/staff/iridis.page}{\textcolor{blue}{https://www.southampton.ac.uk/isolutions/staff/iridis.page}}}, an HPC cluster at the University of Southampton. The experiments were run for 60 hours each. All of the experiments for the heuristic and benchmark algorithms were finished earlier, but some of the runs for CCM\_MADRL\_MEC ran out of time before reaching the last episode. For convenience in plotting with the 95\% confidence interval, all runs of CCM\_MADRL\_MEC are clipped after the run with the least number of episodes. Note that the benchmark and heuristic algorithms have only one Q value in the critic for a combination of state and actions of the actors. On the other hand, the number of Q values to train in the CCM\_MADRL\_MEC is equal to the number of offloaded tasks or 1 if all of them are allocated locally. 

\section{Conclusion}
\label{CCM_MADRL_MEC:Conclution}
In this paper, we propose a combinatorial client-master MADRL algorithm for task offloading in MEC, that considers various constraints at the UDs, sub-channels, and server. By combining the advantages of both policy gradient and value functions to produce continuous and combinatorial actions, CCM\_MADRL provides better convergence than existing homogeneous MADRL algorithms, because the master agent applies combinatorial decisions on the actions proposed by the clients. 

In the future, we plan to extend CCM\_MADRL\_MEC to multi-server MEC where multiple servers cooperate to make combinatorial decisions. 


\begin{acks}
This research was sponsored by the U.S. Army Research
Laboratory and the U.K. Ministry of Defence under Agreement
Number W911NF-16-3-0001. The views and conclusions contained
in this document are those of the authors and should not be
interpreted as representing the official policies, either expressed or implied, of the U.S. Army Research Laboratory, the U.S. Government, the U.K. Ministry of Defence or the U.K. Government. The U.S. and U.K. Governments are authorized to reproduce and distribute reprints for Government purposes notwithstanding any copyright notation hereon. This work was also supported by an EPSRC Turing AI Acceleration Fellowship (EP/V022067/1).

\end{acks}



\bibliographystyle{ACM-Reference-Format} 
\bibliography{AAMAS2024_CCM_MADRL_MEC_arXiv}


\end{document}